\documentclass[10pt, doublecolumn]{IEEEtran}

\usepackage{paralist}
\usepackage[usenames,dvipsnames]{xcolor}
\usepackage{amsthm,amsmath,amsfonts,ed,empheq}
\usepackage{xspace}
\usepackage{url}
\usepackage{multirow}
\usepackage{footnote}
\usepackage{amssymb,amsmath,mathtools,fixmath}
\usepackage{algorithmicx,algorithm}
\usepackage{algpseudocode} 
\usepackage{cases}
\usepackage{cite}
\usepackage{graphicx,amsfonts,psfrag,url,booktabs}
\usepackage{comment}
\usepackage{cite}
\usepackage{caption,subcaption}
\usepackage{makecell}
\usepackage{psfrag}
\usepackage{tabularx}
\usepackage[table]{colortbl}
\usepackage{booktabs}
\usepackage{tikz} 
 
\newcommand*{\Scale}[2][4]{\scalebox{#1}{$#2$}}%

\newcommand{\x}{\mathbold{x}}
\newcommand{\X}{\mathbold{X}}

\newcommand{\y}{\mathbold{y}}
\newcommand{\Y}{\mathbold{Y}}

\newcommand{\w}{\mathbold{w}}
\newcommand{\n}{\mathbold{n}}
\newcommand{\bphi}{\mathbold{\phi}}
\newcommand{\bpsi}{\mathbold{\psi}}

\newcommand{\bPhi}{\mathbold{\Phi}}
\newcommand{\bPsi}{\mathbold{\Psi}}
\renewcommand{\u}{\mathbold{u}}

\newcommand{\I}{\mathbold{I}}
\newcommand{\bSigma}{\mathbold{\Sigma}}
\newcommand{\bLambda}{\mathbold{\Lambda}}
\newcommand{\A}{\mathbold{A}}

\newcommand{\F}{\mathbold{F}}

\newcommand{\eg}{e.g.,\/~}
\newcommand{\ie}{i.e.,\/~}

\newcommand{\wrt}{w.r.t.\xspace}                     

\newcommand{\simG}[1]{\tiny\overset{\mathcal{#1}\vspace{-2mm}}{\sim}}

\newcommand{\dft}[1]{\text{DFT}(#1)}
\newcommand{\gft}[1]{\text{GFT}(#1)}
\newcommand{\jft}[1]{\text{JFT}(#1)}
\newcommand{\idft}[1]{\text{DFT}^{\Scale[0.5]{-1}}(#1)}
\newcommand{\igft}[1]{\text{GFT}^{\Scale[0.5]{-1}}(#1)}
\newcommand{\ijft}[1]{\text{JFT}^{\Scale[0.5]{-1}}(#1)}

\renewcommand{\paragraph}[1]{\vspace{2mm}\noindent \textbf{#1}}

\DeclarePairedDelimiter{\norm}{\lVert}{\rVert}

\DeclareMathOperator*{\argmin}{argmin}

\renewcommand{\L}{\mathbold{L}}

\newcommand{\E}[1]{\mathbf{E}\hspace{-0.7mm}\left[#1\right]}

\newtheoremstyle{slplain}
  {0.7\baselineskip\@plus.2\baselineskip\@minus.2\baselineskip} 
  {0.0\baselineskip\@plus.2\baselineskip\@minus.2\baselineskip}
  {\slshape}{}{\itshape}{.}{ }{}
\theoremstyle{slplain}

\makesavenoteenv{tabular}

\title{Frequency Analysis of Temporal Graph Signals}

\author{Andreas Loukas* and Damien Foucard
\thanks{
The authors are with the Department of Telecommunication Systems, TU Berlin, Germany. e-mails: \{a.loukas, d.foucard\}@tu-berlin.de. *Corresponding author: A. Loukas.}}

\begin{document}

\maketitle

\begin{abstract} 
This letter extends the concept of graph-frequency to graph signals that evolve with time. Our goal is to generalize and, in fact, unify the familiar concepts from time- and graph-frequency analysis. To this end, we study a \emph{joint} temporal and graph Fourier transform (JFT) and demonstrate its attractive properties. We build on our results to create filters which act on the joint (temporal and graph) frequency domain, and show how these can be used to perform interference cancellation. The proposed algorithms are distributed, have linear complexity, and can approximate any desired joint filtering objective.
\end{abstract}


%
%
%
%

\section{Introduction}
\label{sec:intro}

The recent availability of complex and high-dimensional datasets has spurred the need for new data analysis methods. One prominent research direction in signal processing has been the focus on data supported over graphs~\cite{shuman2013}. Graph signals, \ie signals taking values on the nodes of combinatorial graphs, represent a convenient solution to model data exhibiting complex and non-uniform properties, such as those found in social, biological, and transportation networks, among others.
Arguably, the most fundamental tool in the analysis of graph signals is the graph Fourier transform (GFT)~\cite{shuman2012,shuman2013,sandryhaila2014}. 
In an analogous manner to the discrete Fourier transform (DFT), using GFT one may examine graph signals in the graph frequency domain, and, for instance, remove noise by attenuating high graph-frequencies.
GFT has also lead to significant new insights in problems such as smoothing and denoising~\cite{Zhang2008,Shuman2011,Loukas2013}, segmentation~\cite{Loukas2014}, sampling and approximation~\cite{zhu2012,zhu2012b,Narang2013}, and classification~\cite{Smola2003,Zhu2003,Belkin2004} of graph data.  

Yet, for many modern graph datasets, \emph{time} is still of the essence. Whether we are interested in which candidate is more popular to whom in the political blogosphere~\cite{Adamic2005}, how an infection spreads over the global transportation network~\cite{tatem2006}, or what the average daily traffic over the streets of a city is~\cite{mohan2008}, the graph signals one encounters are not only a function of the underlying graph---they also evolve with time. 
Motivated by this need, this paper considers the frequency analysis of graph signals that change with time, referred to as \emph{temporal graph signals}. Our goal is to generalize and, in fact, unify the familiar concepts from time- and graph-frequency analysis so as to jointly consider graph and temporal aspects of data. 

To this end, we advocate for a joint temporal and graph Fourier transform (JFT) constructed by taking the graph Fourier and discrete Fourier transforms jointly. Though this idea is known~\cite{sandryhaila2014}, this paper brings forth new insights: (i) While previously defined only for the adjacency matrix, we show that a joint Fourier transform can be defined over a wider selection of matrix representations of a graph, such as the discrete and normalized Laplacian. (ii) We study the properties and relations between Fourier transforms. This entails providing conditions for JFT to be unitary, as well as showing that JFT and GFT are equivalent under a certain transformation of the input graph, a relation which provides insight into the operation of JFT and demonstrates the consistency of our approach with the established theory. (iii) We propose a generalization of the notion of graph signal smoothness (or more precisely variation) appropriate for temporal graph signals. 


We build on our results to design \emph{joint filters}, which selectively attenuate or amplify certain joint-frequencies of a temporal graph signal. Contrary to previous work~\cite{Loukas2015c,sandryhaila2014}, the proposed filters are distributed, have complexity linear in the number of graph edges, period of the signal and approximation order, and furthermore can approximate \emph{any} desired joint-frequency response. In particular, we demonstrate how they can be used to approximate an interference cancellation problem, where given the statistical properties of a desired and interfering temporal graph signal, one is asked to design the filter which recovers the original signal with the smallest mean-squared error.



\section{Joint Fourier Transform}
\label{sec:JFT}
 
 
\subsection{A Transform for Temporal Graph Signals}
\label{subsec:subsec:jft}

Consider a graph $\mathcal{G} = (V, E)$ of $N$ nodes $u_1, \ldots, u_N$ and $M$ edges and suppose that we are given a periodic temporal graph signal represented by a $\mathbb{R}^{N\times T}$ matrix $\X$, with $X_{nt}$ being the value of node $u_n$ at time instant $t$. Our goal is then to characterize the spectral properties of $\X$. Since each node has as values a temporal periodic signal, common wisdom dictates a transformation from the time to the frequency domain. Applying the discrete-Fourier transform on each row of $\X$, we obtain the frequency representation of our signal
\begin{align}
	\dft{\X} = \X \bPsi_T^\top, 
\end{align}
with the unitary matrix $\bPsi_T = \mathbold{D} / \sqrt{T}$ (i.e., $ \bPsi_T^* \times \bPsi_T = I $) constructed as a normalization of the DFT matrix $\mathbold{D}$ of dimension $T\times T$. However, since the transform matrix $\bPsi_T^\top$ acts on each of the rows of $\X$ \emph{independently}, it overlooks the graph structure of our data. Similarly, applying the graph Fourier transform in parallel~\cite{shuman2013,sandryhaila2014} for each time-instant as 
\begin{align}
	\gft{\X; \mathcal{G}} = \bPsi_G \X ,	
	\label{eq:gft}
\end{align}
where $\bPsi_G$ is the $N\times N$ left eigenvector matrix of a matrix representation of $\mathcal{G}$, such as the Laplacian $\L_G$, normalized Laplacian $\mathbold{N}_G$ or the adjacency matrix $\A_G$, lets us take into account the variation of the signal with respect to the graph, but neglects the temporal aspect of the data.

To capture the frequency content of $\X$ along both temporal and graph domains, one has to apply both transforms \emph{jointly}. We can therefore define a joint graph and temporal Fourier transform as
\begin{align}
	\jft{\X; \mathcal{G}} := \bPsi_G \X \bPsi_T^\top. 
\end{align}
In contrast to~\cite{sandryhaila2014}, the definition above is independent of the matrix representation of $\mathcal{G}$ and can be used in conjuction with each definition of GFT. 
It might be more convenient to express JFT as a matrix vector multiplication. Exploiting the properties of the Kronecker product ($\otimes$), we can write 
\begin{align}
	\jft{\x; \mathcal{G}} = (\bPsi_T \otimes \bPsi_G) \, \x = \bPsi_J \x, 
\end{align}
where in the last step we set $\bPsi_J = \bPsi_T \otimes \bPsi_G$.
%
Let us examine some properties of JFT.

\vskip2mm
\emph{Property 1. JFT is an invertible transform. The inverse transform in matrix and vector form is 
$	\ijft{\Y} = \bPsi_G^{-1} \Y \bPsi_T^{\top^*} \quad \text{and} \quad \ijft{\y} = (\bPsi_T^* \otimes \bPsi_G^{-1})\, \y, \notag 
$
respectively, where $\y = \text{vec}(\Y) = \jft{\x}$.} This can be confirmed using the fact that $\bPsi_T$ is unitary. For convenience of notation, in the following we set $\bPhi_G = \bPsi_G^{-1}$, $\bPhi_T = \bPsi_T^*$ and $\bPhi_J = \bPhi_T \otimes \bPhi_G$, such that $\idft{\y} = \bPhi_T \y$, $\igft{\y} = \bPhi_G \y$, and $\ijft{\y} = \bPhi \y$.

\vskip2mm
\emph{Property 2. JFT is a unitary transform if and only if GFT is unitary. } 
JFT is a unitary transform when $\bPsi_J \bPsi_J^* = \I_{NT}$. From definition, we have
\begin{align}
	\bPsi_J \bPsi_J^* &= (\bPsi_T \otimes \bPsi_G) (\bPsi_T \otimes \bPsi_G)^* \nonumber \\
	&= (\bPsi_T \bPsi_T^*) \otimes (\bPsi_G \bPsi_G^*) = \I_T \otimes (\bPsi_G \bPsi_G^*).
\end{align}
For the last statement to be equal to $\I_{NT}$ (\ie an identity matrix of dimension $NT$) it must be that $\bPsi_G \bPsi_G^* = \I_N$, which is equivalent to asserting that GFT is unitary. 

We deduce that JFT is a unitary transform for all symmetric matrix representations of a graph, such as the Laplacian or adjacency matrix, as long as the graph is undirected. On the other hand, when the graph is directed, unitarity is lost. 
It also follows that, if a symmetric matrix representation is used: (\emph{i}) The columns of $\bPhi_J$ form an orthonormal basis, and (\emph{ii}) JFT obeys the Parseval theorem.
For clarity, in the rest of this paper we opt to work with undirected graph $\mathcal{G}$ and only express our results \wrt the Laplacian matrix. Still, all results are directly applicable to alternative matrix representations.


\vskip2mm
\emph{Property 3. JFT is independent of the order DFT and GFT are applied on $\X$}. 
This is a direct consequence of the associativity of matrix multiplication.


\subsection{Transform Equivalence}

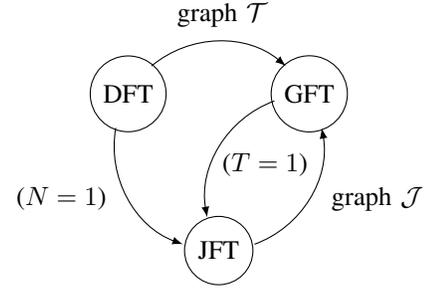
\begin{figure}
\begin{center}
\vspace{-2mm}\hspace{10mm}
\begin{tikzpicture}
\def \n {3}
\def \radius {1.39cm}
\def \margin {20} 
\def \offset {150}

  \node[draw, circle] at ({360/\n * (1 - 1)+\offset}:\radius) {DFT};
  \node[draw, circle] at ({360/\n * (2 - 1)+\offset}:\radius) {JFT};
  \node[draw, circle] at ({360/\n * (3 - 1)+\offset}:\radius) {GFT};

  \draw[->, >=latex] ({360/\n * (1 - 1)+\margin+\offset}:\radius) 
    arc ({360/\n * (1 - 1)+\margin+\offset}:{360/\n * (1)-\margin+\offset}:\radius);
 
    
  \draw[<-, >=latex] ({360/\n * (3 - 1)+\margin+\offset}:\radius) 
    arc ({360/\n * (3 - 1)+\margin+\offset}:{360/\n * (3)-\margin+\offset}:\radius);

  \draw[->, >=latex] ({360/\n * (2 - 1)+\margin+\offset}:\radius) 
    arc ({360/\n * (2 - 1)+\margin+\offset}:{360/\n * (3-1)-\margin+\offset}:\radius);

  \draw[->, >=latex] (0.74cm,0.6cm) 
    arc ({360/\n * (2 - 1)+\margin+\offset-180}:{360/\n * (3-1)-\margin+\offset-180}:\radius);

\node[text width=2cm] at (0.45cm,1.75cm) {graph $\mathcal{T}$};
\node[text width=2cm] at (-1.69cm,-0.7cm) {($N=1$)};
\node[text width=2cm] at (2.5cm,-0.7cm) { graph $\mathcal{J}$};
\node[text width=2cm] at (1.05cm,-0.25cm) {($T=1$)};
\end{tikzpicture}
\end{center}\vspace{-2mm}
\caption{\small Relations between Fourier transforms. Each directed arrow (say from $A$ to $B$) in the figure should be interpreted as a transform-simulation (transform A can be simulated by B). Edge annotations hint on the simulation method.\vspace{-4mm}}
\label{fig:relations}
\end{figure}

Suppose that we are given a graph $\mathcal{G}$ of $N$ nodes and a periodic temporal graph signal $\X$ of period $T$. 
Fig.~\ref{fig:relations} characterizes the relations between DFT, GFT, and JFT of $\X$. Each directed arrow (\eg from $A$ to $B$) in the figure should be interpreted as a transform-simulation (transform A can be simulated by B). The equivalence between GFT and JFT is illustrated as a bidirectional simulation. 
Let us begin from the obvious relations. By definition, both DFT and GFT are specific cases of JFT. In particular, $\dft{\X} = \jft{\X;\mathcal{G}}$ if the graph consists of a single node ($N = 1$) and $\gft{\X;\mathcal{G}} = \jft{\X;\mathcal{G}}$ if $\X$ does not change in the temporal domain ($T=1$). 
	
We proceed with the remaining two relations.


%

\vskip1mm\emph{DFT $\rightarrow$ GFT}. To establish that DFT can be simulated using GFT, we will identify a graph $\mathcal{T} = (V_{T}, E_{T})$ such that $\dft{\X} = \gft{\X;\mathcal{T}}$. The last relation is equivalent to requiring that the left and right eigenvector matrices of the Laplacian $\L_T$ of graph $T$ are $\bPsi_T$ and $\bPhi_T$, respectively. 
We obtain $\mathcal{T}$ by thinking of (periodic) time as a ring graph consisting of $T$ nodes, one per time-instant~\cite{Puschel2008,sandryhaila2013}. In other words, each node $u_t \in V_{T}$ is connected to node $u_{t+1}$ for $t = 1,\ldots,T$, with index $T+1 = 1$. The adjacency matrix of $\mathcal{T}$
is a circulant matrix and is known to have $\bPsi_T$ and $\bPhi_T$ as left and right eigenvector matrices, and as eigenvalues $\lambda_T(t) = \text{exp}({(2\pi i (t-1)(T-1)) / T })$.
Furthermore, since the Laplacian of $\mathcal{T}$ is given by $\L_T = \I_T - \A_T$, it has the same eigenvectors and eigenvalues (up to translation and reordering), rendering the choice of representation (between $\A_T$ or $\L_T$ or $\mathbold{N}_T$) arbitrary. 

\begin{figure}[t]
	\centering
	\hspace{-0mm}
	\includegraphics[width=1.\columnwidth]{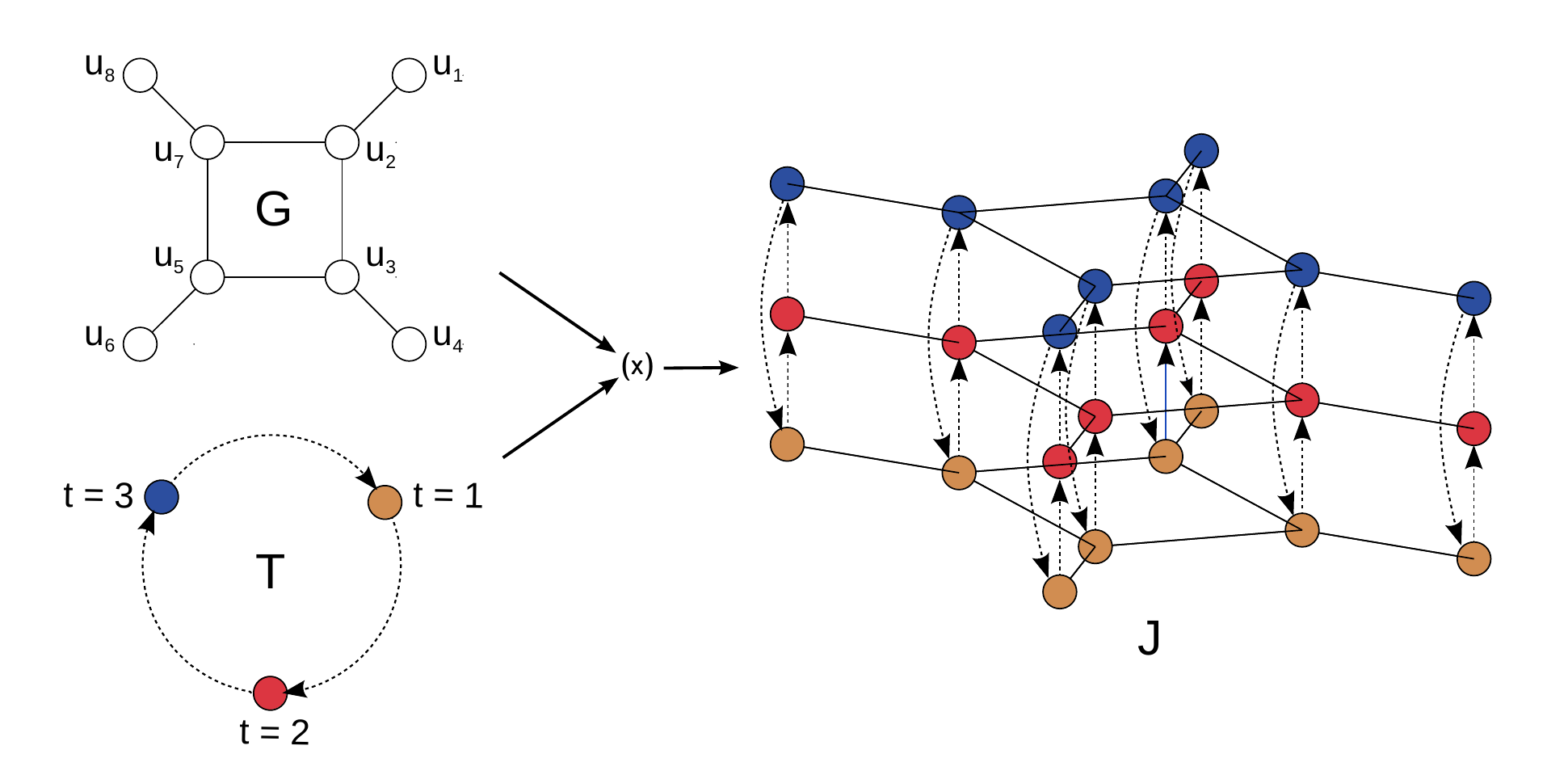}
	\vspace{-4mm}
	\caption{The joint graph $\mathcal{J}$ is the graph cartesian product of the input graph $\mathcal{G}$ and a cycle graph $\mathcal{T}$.\vspace{-4mm}}
	\label{fig:jointGraph}
\end{figure}

\vskip1mm\emph{JFT $\rightarrow$ GFT.} We will simulate JFT by applying GFT on the joint graph $\mathcal{J} = (V_J, E_J)$, effectively showing that $\jft{\X;\mathcal{G}} = \gft{\X;\mathcal{J}}$.
We construct $\mathcal{J}$ as the graph cartesian product of $\mathcal{G}$ and $\mathcal{T}$. The joint graph consists of $T$ copies of $\mathcal{G}$, denoted by $\mathcal{G}_{t} = (V_{t}, E_{t})$, one for each time-instant, with $V_J = V_1\cup \ldots \cup V_T$. Name the corresponding nodes in each copy as $u_{n,t} \in V_{t}$. In addition to the $T \times M$ edges already introduced, the joint graph contains $T\times N$ extra edges joining consecutive copies: in particular, for each node $u_{n,t}$ in $\mathcal{G}_{t}$ the joint graph has a \emph{directed} edge to node $u_{n,t+1}$ in $\mathcal{G}_{t+1}$ (modulo $T$). The Laplacian matrix\footnote{The argument is identical for the adjacency and normalized Laplacian matrix representations.} of $\mathcal{J}$ is expressed as 
\begin{align}
	\L_J &= \I_T \otimes \L_G + \L_T \otimes \I_N = \L_T \oplus \L_G,
\end{align}
where $(\oplus)$ is the knonecker sum operator. Even though $\L_J$ is not a symmetric matrix (due to $\mathcal{T}$ and $\mathcal{J}$ being directed) it follows from Theorem~13.16 in~\cite{Laub2005} that $\L_J$ has eigendecomposition
\begin{align}
	\L_J &= (\bPhi_T\otimes\bPhi_G) (\bLambda_T \oplus \bLambda_G) (\bPsi_T\otimes\bPsi_G) = \bPhi_J \bLambda_J \bPsi_J, \notag
\end{align}
which fulfills out requirement.

\subsection{A Joint Notion of Variation} 
\label{subsec:variation}

The utility of a transform stems largely from its ability to provide insight about data. For instance, by observing the GFT of a graph signal one gains intuition about the \emph{variation} of a signal over the graph, a notion which 
characterizes how aggressively a signal is changing on the graph. Therefore, GFT is useful because it allows us to distinguish smooth signals from non-smooth ones. 
In a similar manner, to render JFT a useful transform, beyond that of being a combination of two other transforms, we must give it insightful meaning. 

We propose to use the relation $\jft{\X;\mathcal{G}} = \gft{\X;\mathcal{J}}$ in order to imbue JFT with an appropriate notion of smoothness. In this way the variation of a temporal graph signal $\X$ on $\mathcal{G}$ is defined to be equal to the variation of the same signal (interpreted now as a graph signal) on the joint graph $\mathcal{J}$. 

Consider a node $u_{n_i,t} \in V_t \subseteq V_J$ and denote by $n_j \simG{J} n_i$ its neighbors in $\mathcal{J}$. Using the definition of variation~\cite{shuman2013} on $\mathcal{J}$, we define the \emph{local variation} of the temporal graph signal $\X$ at the $n_i$-th node at time $t$ to be \vspace{-4mm}
\begin{align}
	\norm{\nabla_{n_i,t} \X}_2 &:= \left[ \sum_{n_j \simG{J} n_i} \left( \frac{\partial \X}{ \partial e_{n_i n_j}}  \right)^2 \right]^{\frac{1}{2}} \notag \\ 
	&\hspace{-20mm}= \left[ \sum_{n_j \simG{G} n_i} \hspace{-2mm} \left(X_{n_j,t} - X_{n_i,t} \right)^2 + \left(X_{n_i,t-1} - X_{n_i,t}\right)^2 \right]^{\frac{1}{2}}\hspace{-2mm},
\end{align}
where $\frac{\partial \X}{ \partial e_{n_i n_j}}$ is the discrete edge derivative on the joint graph and, in the last equation, $n_j \simG{G} n_i$ are the neighbors of $u_{n_i}$ in $\mathcal{G}$. 
We can also obtain a global notion of smoothness using the $p$-Dirichlet form
\begin{align}
	S_p(\X) &:= \frac{1}{p} \sum_{n=1}^N \sum_{t = 1}^p \norm{\nabla_{n,t} \X}_2^p \\
		&\hspace{-12mm}= \frac{1}{p} \sum_{n_i = 1}^N \hspace{-1mm}\left[ \sum_{n_j \simG{G} n_i} \hspace{-2mm} \left(X_{n_j,t} - X_{n_i,t} \right)^{\hspace{-0.5mm}2} \hspace{-0.5mm}+ \hspace{-0.5mm} \left(X_{n_i,t-1} - X_{n_i,t}\right)^2 \right]^{\hspace{-1mm}\frac{p}{2}}\hspace{-2mm}\notag\hspace{0mm}.
\end{align}
For $p = 2$ and after some manipulation, we find that
\begin{align}
	S_2(\X) 
	&= \text{vec}(\X)^\top \L_J \text{vec}(\X) = \x^\top \L_J \x.
\end{align}
Similarly to the GFT, $S_2(\X)$ is a quadratic form of the (joint) Laplacian, which implies that $S_2(\X)\geq 0$. Yet, here the variation of a signal is not only \wrt $\mathcal{G}$ but also \wrt time. For instance, $S_2(\X) = 0$ only if the signal is constant across all nodes and time-instances and, in general, the slower the values change along the graph and temporal domains, the smaller $S_2(\X)$ becomes. Moreover, according to the Courant-Fischer theorem, the signals which minimize $S_2(\X)$ are exactly the eigenvectors of $\L_J$ (\ie the rows of $\bPsi_J$) with the corresponding minima being the associated eigenvalues $\lambda_J$ of $\L_J$~\cite{Horn2012}. JFT therefore characterizes a signal by how close its projections lie to the minimizers of the global variation $S_2(\X)$; meaning that terms of low joint frequency $\lambda_J$ (projections to eigenvectors associated with small eigenvalues) correspond to smoother signals and vice-versa. 


\section{Joint Filters}
\label{sec:filters}



To define the joint filtering problem in the most general form, 
we will consider a two-dimensional frequency domain, with the two dimensions conveying respectively the time- and graph- frequency of the signal. That is, one may now define joint frequency response $h(\lambda_T, \lambda_G)$ describing how the filter should change the frequency components independently \wrt $\lambda_T$ and $\lambda_G$. 
\begin{align}
	\F \x = \hspace{-1mm}\sum_{t = 1, n = 1}^{T,N} \hspace{-2mm} h(\lambda_{T}(t), \lambda_{G}(n)) \, \bphi_{J}(t,n) \bpsi_{J}^*(t,n) \, \x
	\label{eq:jointFilter2D}
\end{align}
Above $\bpsi^*_{J}(t,n) = \bpsi^*_{T}(t) \otimes \bpsi^*_{G}(n)$ and $\bphi_{J}(t,n) = \bphi_{T}(t) \otimes \bphi_{G}(n)$ are the eigenvectors giving eigenvalue $\lambda_{T}(t) + \lambda_{G}(n)$. 

\begin{figure}[t]
\centering
    \begin{subfigure}[b]{0.49\columnwidth}
        \includegraphics[width=1\columnwidth]{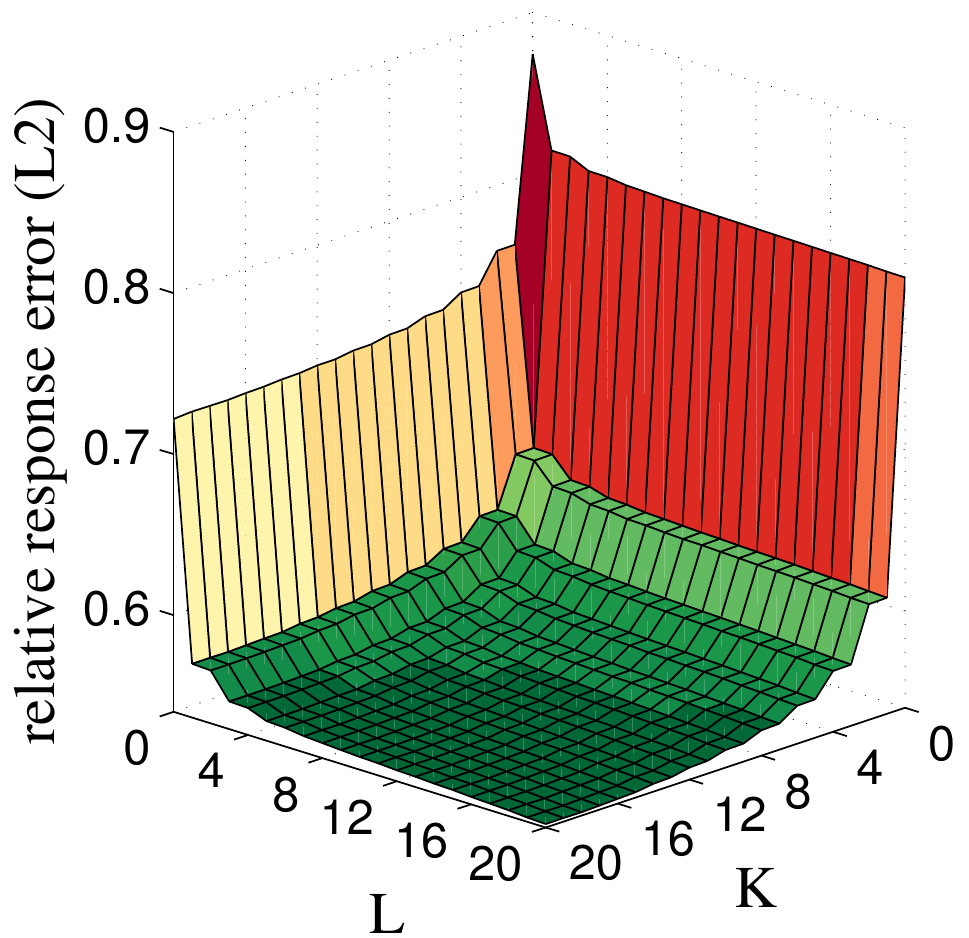}
        \caption{ideal low-pass}
        \label{subfig:step}
    \end{subfigure}
    \begin{subfigure}[b]{0.49\columnwidth}
        \includegraphics[width=1\columnwidth]{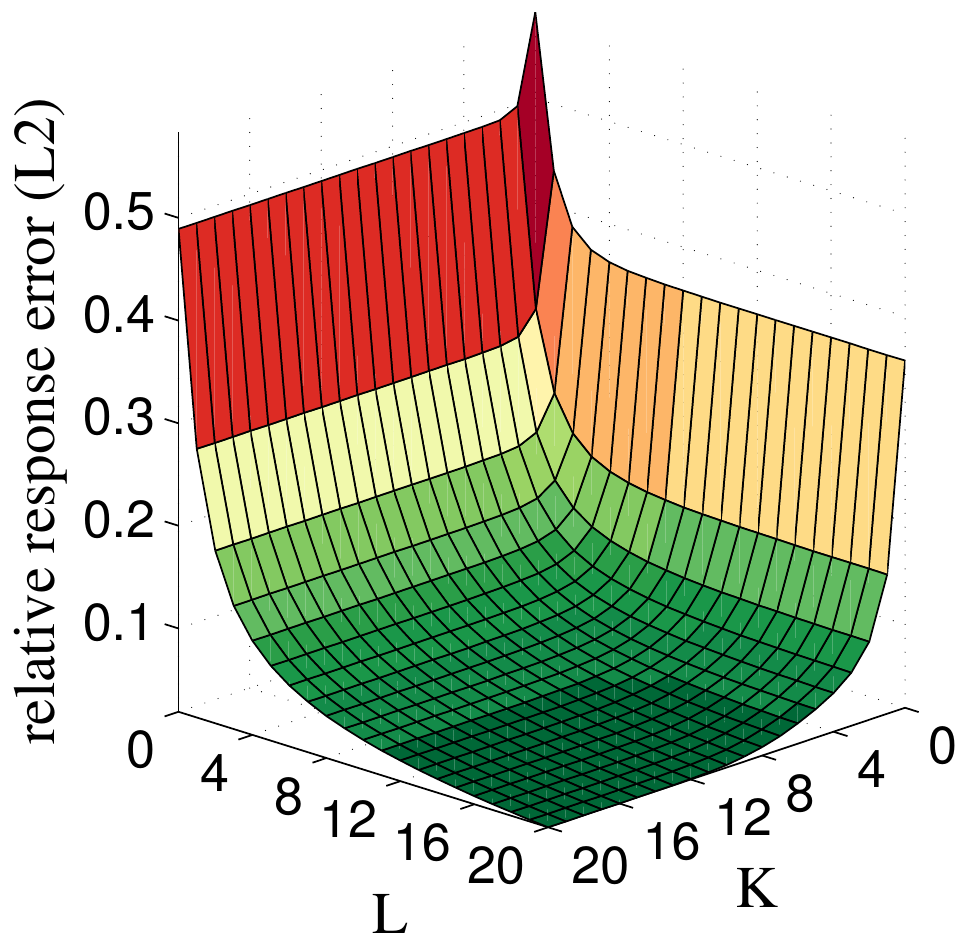}
        \caption{interference cancellation}
        \label{subfig:denoising}
    \end{subfigure}    
    \caption{\small Goodness of joint polynomial approximation of (a) an ideal joint low-pass filter and (b) a rational function used for interference cancellation, for varying approximation orders. \vspace{-4mm}}%
    \label{fig:goodJointFilterApproxStep}
\end{figure}

\paragraph{Joint filter design.} In the spirit of FIR filters~\cite{Shuman2011}, our approach will be to first approximate the desired joint frequency response $h^\star(\lambda_T, \lambda_G)$ by a bivariate polynomial of orders $K,L$ in time and graph, respectively
\begin{align}
	h(\lambda_T, \lambda_G) = \sum_{k = 0, \ell = 0}^{K,L} c_{kl} \lambda_T^k \lambda_G^\ell,
\end{align}
with coefficients $c_{kl}$ chosen to minimize a certain norm (such as the max-norm or the euclidean norm) of the approximation error.  
The corresponding joint filter is 
\begin{align}
	\F \x &= \sum_{k = 0, \ell = 0}^{K,L} c_{kl} \, (\L_T^k \otimes \L_G^\ell) \, \x
 \label{eq:jFIR}
\end{align}
which, can be easily shown to possess the required response (by taking the eigenvalue decomposition of $\L_T$ and $\L_G$, applying the mixed-product property of the Kronecker product, and exchanging the sums).
%
%
Figure~\ref{subfig:step} illustrates the approximation error $\norm{h^\star - h}_2/\norm{h^\star}_2$ in the challenging case when the desired response is an ideal low-pass filter in time as well as graph frequency, $h^\star(\lambda_T, \lambda_G) = 1$ if $ \text{angle}(\lambda_T) \leq \pi $ and $ \lambda_G \le 1$, and $h^\star(\lambda_T, \lambda_G) = 0$, otherwise (note that since we use the normalized Laplacian, $0 \leq \lambda_G \leq 2$). Observe that the quality of approximation increases with the polynomial degrees, leading to a error of 0.54 for $K,L \geq 15$. Smaller errors can be achieved for continuous $h^\star$ functions (see for instance Fig.~\ref{subfig:denoising}). 

\emph{Computation.} Because~\eqref{eq:jFIR} involves powers of the input signal, it can be computed distributedly: each term $\L_{T}^k \otimes \L_G^\ell \x $ is computed by iteratively multiplying the signal by $\I_N\otimes \L_G$ ($\ell$ times) and $\L_T\otimes \I_N$ ($k$ times), with each multiplication being a local operator on the joint graph and requiring the communication of $2MT$ and $NT$ values, respectively.
We can also reduce the overall complexity by computing terms $\L_{T}^k \otimes \L_G^\ell \x $ recursively from either $\L_{T}^k \otimes \L_G^{\ell-1} \x $ or $\L_{T}^{k-1} \otimes \L_G^{\ell} \x$.
Since $2MT \geq NT$, the most efficient scheme, which involves computing first all powers $\I_T \otimes \L_G^\ell \x $ and then using them to compute remaining terms, requires the exchange of $2MTK + (K+1)NTL = TK(2M + NL) + NTL = O(MTKL)$ values.




\paragraph{Interference cancellation.} Suppose that we want to recover a graph signal $\u \in \mathbb{R}^{NT}$ from an interfering signal $\w \in \mathbb{R}^{NT}$. 
In this problem instance however, the two signals possess strong statistical structure in different domains. Let $\mathbold{f}$ and $\mathbold{g}$ be arbitrary matrix functions. We assume that, whereas $\u$, which has covariance $\bSigma_u = \I_T \otimes \mathbold{g}(\L_G)$, has statistical properties that are \emph{not} a function of time, the opposite holds for $\w$ with covariance $\bSigma_w = \mathbold{f}(\L_T) \otimes \I_N $, whose statistical properties are entirely temporal (this is a generalization of the models in~\cite{Zhang2015}). 

For simplicity, suppose that both signals are zero-mean. According to Wiener filter theory, the linear operator $\bar{\F}$ that recovers $\u$ from $\x = \u + \w$ with minimal mean-squared error is
\begin{align}
\hspace{-3mm}	\bar{\F} \hspace{-0.5mm} = \hspace{-0.5mm}\argmin_{\F} \E{ \frac{\norm{ \F\x  - \u}_2^2}{NT} } \hspace{-0.5mm}= \hspace{-0.5mm}\bSigma_u ( \bSigma_u + \bSigma_w)^{\dagger}, 
\end{align}
where using the pseudo-inverse (${\dagger}$) instead of the normal matrix inverse allows us to extend the result to positive semi-definite covariances. We then have that
\begin{align}
	(\bSigma_u + \bSigma_w)^{\dagger} &= (\I_T \otimes \mathbold{g}(\L_G) + \mathbold{f}(\L_T) \otimes \I_N)^{\dagger} \nonumber \\
						  &= (\mathbold{f}(\L_T) \oplus \mathbold{g}(\L_G))^{\dagger} \notag \\
						  &= \bPhi_J \, \left(\mathbold{f}(\bLambda_T) \oplus \mathbold{g}(\bLambda_G)\right)^{\dagger} \, \bPsi_J.
\end{align}
%
%
In addition, 
\begin{align}
	\bSigma_u = \I_T \otimes \mathbold{g}(\L_G) &= \bPhi_T \bPsi_T \otimes \bPhi_G \mathbold{g}(\bLambda_G) \bPsi_G \nonumber \\
	&\hspace{-10mm}= (\bPhi_T \otimes \bPhi_G) (\I_T \otimes \mathbold{g}(\bLambda_G)) (\bPsi_T  \otimes \bPsi_G  ) \nonumber \\
	&\hspace{-10mm}= \bPhi_J \, (\I_T \otimes \mathbold{g}(\bLambda_G)) \, \bPsi_J.
\end{align}
Since $\bPsi_J \bPhi_J = \I_{NT}$, we conclude that
%
%
%
\begin{align}
\hspace{-2mm}\bar{\F} = \bPhi_J \, (\I_T \otimes \mathbold{g}(\bLambda_G)) \left(\mathbold{f}(\bLambda_T) \oplus \mathbold{g}(\bLambda_G)\right)^{\dagger} \, \bPsi_J,
\end{align}
which is a joint filter with response 
$h(\lambda_T, \lambda_G) = g(\lambda_G) / (g(\lambda_G) + f(\lambda_T))$ if $g(\lambda_G) + f(\lambda_T) \neq 0$ and $h(\lambda_T, \lambda_G) = 0$, otherwise.
Therefore, signals $\u$ and $\w$  cannot be well separated by acting disjointly on their respective domains; the best linear estimator $\bar{\F} \x$ of $\u$, is given by a filter acting on the joint Fourier domain. Moreover, as shown in Fig.~\ref{subfig:denoising} (now for the simple case of $f(\lambda_T) = \lambda_T + 1$ and $g(\lambda_G) = \lambda_G$) operator $\bar{\F}$ can be tightly approximated by our proposed joint filters~\eqref{eq:jFIR}, even using moderate polynomial orders.

\bibliographystyle{IEEEtran}
\bibliography{bibliography.bib}

\end{document}